\documentclass[10pt,twocolumn,letterpaper]{article}

\usepackage{wacv}
\usepackage{times}
\usepackage{epsfig}
\usepackage{graphicx}
\usepackage{amsmath}
\usepackage{amssymb}
\makeatletter
\@namedef{ver@everyshi.sty}{}
\makeatother

\usepackage{balance}       
\usepackage{graphics}      
\usepackage[T1]{fontenc}   
\usepackage{txfonts}
\usepackage{mathptmx}
\usepackage[pdflang={en-US},pdftex]{hyperref}
\usepackage{color}
\usepackage{booktabs}
\usepackage{textcomp}
\usepackage{microtype}        
\usepackage{ccicons}          

\usepackage{todonotes}

\usepackage{booktabs} 
\usepackage{xcolor} 
\usepackage{eso-pic}
\usepackage{xspace}
\usepackage{times}
\usepackage{float}
\usepackage{tcolorbox}
\usepackage{epsfig}
\usepackage{graphicx}
\usepackage{amsmath}
\usepackage{amssymb}
\usepackage{color}
\usepackage{url}
\usepackage{tabu}
\usepackage{pifont}
\usepackage{pgfplots}
\usepackage{arydshln}
\usepackage{adjustbox}
\usepgfplotslibrary{external}
\usetikzlibrary{positioning}
\graphicspath{{./images/}}
\usepackage{array,multirow,graphicx}
\usepackage{sidecap}
\usepackage{wrapfig}
\usepackage{pifont}
\newcommand{\cmark}{\ding{51}}%
\newcommand{\xmark}{\ding{55}}%

\usepackage{etoolbox}

\wacvfinalcopy 

\def\wacvPaperID{341} 

\ifwacvfinal\fi
\begin{document}
\title{Object Referring in Visual Scene with Spoken Language}

\author{Arun Balajee Vasudevan$^{1}$, Dengxin Dai$^{1}$, Luc Van Gool$^{1,2}$  \\
ETH Zurich$^{1}$ \hspace{1.5cm} KU Leuven $^{2}$  \\
{\tt\small \{arunv,dai,vangool\}@vision.ee.ethz.ch}
     \thanks{$^{1}$Arun Balajee Vasudevan, Dengxin Dai, and Luc Van Gool are with the Toyota TRACE-Zurich team at the Computer Vision Lab, 
        ETH Zurich, 8092 Zurich, Switzerland}%
\thanks{$^{2}$Luc Van Gool is also with the Toyota TRACE-Leuven team at the Dept of Electrical Engineering ESAT, KU Leuven
         3001 Leuven, Belgium}%
}

\maketitle
\ifwacvfinal\thispagestyle{empty}\fi
\begin{abstract}
Object referring has important applications, especially for human-machine interaction. While having received great attention, the task is mainly attacked with written language (text) as input rather than spoken language (speech), which is more natural. This paper investigates Object Referring with Spoken Language (ORSpoken) by presenting two datasets and one novel approach. Objects are annotated with their locations in images, text descriptions and speech descriptions. This makes the datasets ideal for multi-modality learning. The approach is developed by carefully taking down ORSpoken problem into three sub-problems and introducing task-specific vision-language interactions at the corresponding levels. Experiments show that our method outperforms competing methods consistently and significantly. The approach is also evaluated in the presence of audio noise, showing the efficacy of the proposed vision-language interaction methods in counteracting background noise. 
\end{abstract}

\section{Introduction}
\label{sec:intro} 
The recent years have witnessed a great advancement in the subfields of Artificial Intelligence~\cite{lecun2015deep}, such as Computer Vision, Natural Language Processing and Speech Recognition. The success obtained in these individual fields
necessitates formulating and addressing more AI-complete research problems.  This is evidenced by the recent trend of a joint understanding of vision and language in tasks such as image/video captioning~\cite{show:tell:caption}, visual question answering~\cite{VQA}, and object referring~\cite{mao2016generation}. This work addresses the task of object referring (OR). 

OR is heavily used in human communication; the speaker issues a referring expression; the co-observers then identify the referred object and continue the dialog. Future AI machines, such as cognitive robots and autonomous cars, are expected to have the same capacities for effective human-machine interaction. OR has received increasing attention in the last years with large-scale datasets compiled~\cite{kazemzadeh2014referitgame, mao2016generation, yu2016modeling} and sophisticated learning approaches developed~\cite{mao2016generation,hu2016natural,Nagaraja2016, phloc,wang2016structured}. Despite the progress, OR has been mainly tackled with clean, carefully-written texts as inputs  rather than with speech. This requirement largely hinders the deployment of OR to real applications such as assistive robots and automated cars, as we human express ourselves more naturally in speech than in writing. This work addresses object referring with spoken languages (ORSpoken) in constrained condition, in the hope to inspire research under more realistic setting. 

\begin{figure*}[!bt]
\centering
\adjustbox{trim={.00\width} {.0\height} {0.00\width} {.0\height},clip}{
\includegraphics[width=0.95\textwidth]{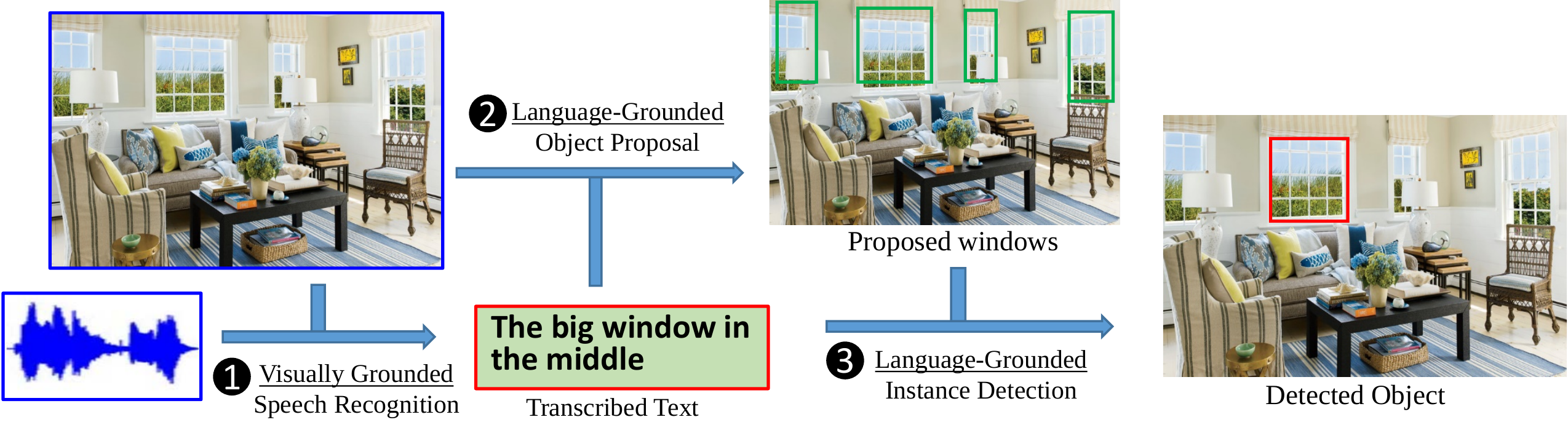}}
\caption{The illustration of our approach for Object Referring with Spoken Language. Given an image and the speech expression, our approach first transcribes the speech via its Visually Grounded Speech Recognition module. It then proposes class-specific candidates by its Language-Grounded Object Proposal module. Finally, the method identifies the referred object via its Language-Grounded Instance Detection module. The inputs of our method are marked in blue and the outputs in red.}
\label{fig:full-pipeline}
\end{figure*}

This work makes two main contributions: \textbf{(i)} we compile \textbf{two datasets} for ORSpoken, one for assistive robots and the other for automated cars; and \textbf{(ii)} we develop \textbf{a novel approach} for ORSpoken, by carefully taking  the problem down into three sub-problems and introducing goal-directed interactions between vision and language therein.  All data are manually annotated with location of objects in images, and text and speech descriptions of the objects. This makes the datasets ideal for research on learning with multiple modalities. Coming to the approach, ORSpoken is decomposed into three sub-tasks: Speech Recognition (SR) to transcribe speech to texts,  Object Proposal to propose candidates of the referred object, and Instance Detection to identify the referred target out of all proposal candidates. The choice of this architecture is mainly to: 1) better use the existing resources in these `subfields';  and 2) better explore multi-modality information with the simplified learning goals. The pipeline of our approach is sketched in Figure~\ref{fig:full-pipeline}. 

More specifically, we ground SR to the visual context for accurate speech transcription as speech content and image content are often correlated. For instance, transcription \emph{`window ...'} is more sensible than \emph{`widow ...'} in the visual context of a living room. For Object Proposal, we ground it to the transcribed text to mainly propose candidates from the referred object class. For instance, to  propose windows, instead of objects of other classes such as sofa, for the expression \emph{`the big window in the middle'} in Figure~\ref{fig:full-pipeline}. Lastly, for Instance Detection, an elaborate model is learned to identify the referred object from the proposed object candidates. For example, the middle window out of the four proposed windows in Figure~\ref{fig:full-pipeline}.

We evaluate the method on the two compiled datasets. Experiments show that the proposed method outperforms competing methods consistently for both expression transcription and for object localization. This work also studies the performance of the method in the presence of different levels of audio noise. The introduced modules for language-vision interaction are found especially useful for large noise. These  interaction components  are general and can be applied to other relevant tasks as well, such as speech-based visual question answering, and speech-based image captioning. 

\section{Related Work}

\noindent
\textbf{Language-based Object Detection}.
Language-based object detection (LOD) has been tackled under different names in Computer Vision. 
Notable ones are referring expressions \cite{yu2016modeling,mao2016generation}, phrase localization \cite{phloc,wang2016structured}, grounding of textual phrases \cite{rohrbach2016grounding}, language-based object retrieval~\cite{hu2016natural} and segmentation~\cite{hu2016segmentation}. 
Recent research focus on LOD  can be put into 2 groups: 1) learning embedding functions~\cite{multimodal:pooling,deep:semantic:alignment:cvpr15} for effective interaction between vision and language; 2) modeling contextual information to better understand a speaker's intent, be it global context~\cite{mao2016generation,hu2016natural}, or local among `similar' objects~\cite{Nagaraja2016,yu2016modeling,mao2016generation}.  Our work use natural speech rather than clean texts as the input. 

Similar work has been conducted in Robotics Community. For instance, a joint model is learned  from gestures, languages, and visual scenes for human-robot interaction in~\cite{human:robot:interaction:aaai:14} ; different network architectures have been investigated for interpreting contextually grounded natural language commands \cite{natural:language:communication:robot:16}; and  the  appropriateness  of  natural  language  dialogue  with assistive robots is examined in \cite{natural:dialogue:robot:06}. While sharing similarity, our work are  very different. Our method works with wild, natural scenes, instead of   rather controlled  lab environments.




\noindent
\textbf{Joint Speech-Visual Analysis}.
Our work also shares similarity with visually grounded speech understanding. Notable examples include visually-grounded instruction understanding~\cite{grounding:verbs:14, tell:me:dave}, audio-visual speech recognition~\cite{audio:visual:SR}, and a `unsupervised' speech understanding via a joint embedding with images~\cite{speech:caption,visual-grounded:speech:17}. 
The first vein of research investigates heavily how to gain the benefit of visual cues to ground high-level commands to low-level actions. The second stream is to use image processing capabilities in lip reading to aid speech recognition systems. The last school aims to reduce the amount of supervision needed for speech understanding by projecting spoken utterances and images to a joint semantic space.  Our work exploits the visual context/environment for better instruction recognition.   


\noindent
\textbf{Speech-based User Interface}.
Speech is a primary means for human communication. Speech-based User Interface has received great attention in the last years. This is evidenced by the surge of commercial products such as Apple's Siri, Microsoft's Cortana, Amazon's Echo, and Google Now,  and a large body of academic publications~\cite{role:hmc:95, avicar:04, wired4speech, Qme, jacko2012human, in-vehicle:dialog}. 
Speaking has been proven faster than typing on mobile devices for exchanging information~\cite{speech:faster}.  There are academic works~\cite{show:tell,   pixel:tone, image:spirit, dai:phd:thesis} which use speech for image description and annotation.  The main merit is that speech is very natural and capable of conveying rich content, lively emotion, and human intelligence -- all in a hands-free and eyes-free manner. Our work enhances Speech-based User Interface with better object localization capability. 

\section{Approach}
\label{sec:approach}
Given an image and a speech expression for the target object, the goal of our method is 1) to transcribe the speech expression into text and 2) to localize the referred object in the image with an axis-aligned bounding box.  As described, our method consists of three components: a) Visually Grounded Speech Recognition(VGSR) for transcribing speech to text, b) Language grounded Object Proposal (LOP) to propose object candidates, and c) Language-Grounded Instance Detection (LID) (Section~\ref{subsection:lod}) for localizing the specific object instance.  See Figure~\ref{fig:full-pipeline} for an illustration .   

\begin{figure*}[tb]
\centering
\includegraphics[width=1\textwidth]{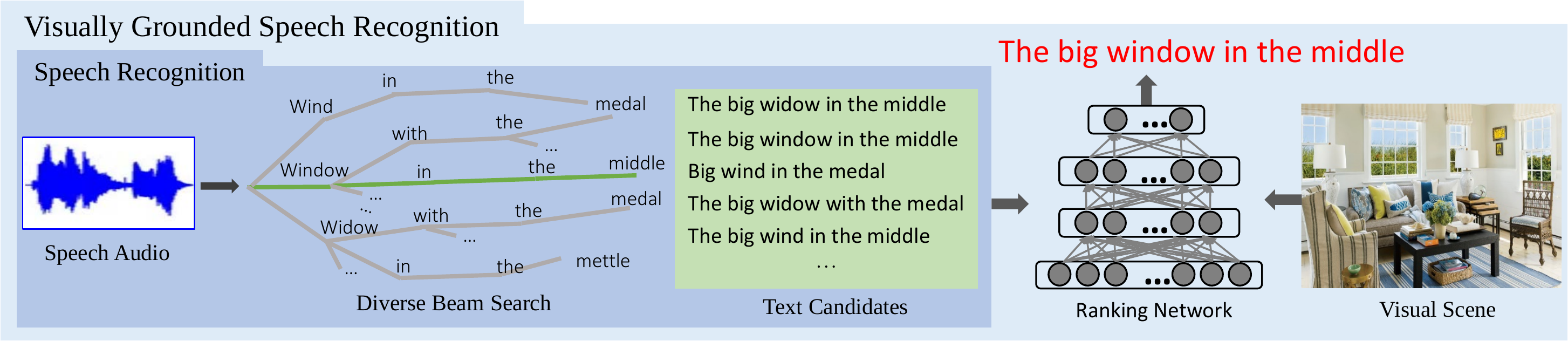}
\vspace{-3mm}
\caption{The pipeline of Visually Grounded Speech Recognition (VGSR). Given an image and a speech, we output transcribed text for the same speech using the image as context. We use Google API to yield a diverse set of transcribed texts. Later, a ranking model to score each of the text given the contextual features from the image and choose the highly ranked one.}
\label{fig:pipeline}
\vspace{-4mm}
\end{figure*}

\subsection{Visually grounded Speech Recognition(VGSR)}
\label{subsection:vgsr}
Given a human generated speech and the context of speech in the form of an image, the main objective of this part is to output the transcribed text for the given speech.
In order to better show insights, we keep the method simple in this work by adopting a holistic ranking approach. In particular, given an image and a speech data referring an object, a list of alternatives for transcribed text are generated by a speech recognition(SR) engine. We then learn a joint vision-language model to choose the most sensible one based on the visual context. 

For a given speech input, the SR engine yields a diverse set of transcribed texts  by Diverse Beam Search as shown in Figure~\ref{fig:pipeline}. Here, speech recognition results are proposed agnostic to the visual context. Nevertheless, we have a prior information that speech refers to an object in the image. Hence, we learn a similarity model to rank the speech recognition results conditioned on the image. 
The model is learned to score the set of transcribed alternatives using a regression loss. The inputs of the model are the embedded representations of image and text.

We use Google Cloud Speech API service~\footnote{\url{https://cloud.google.com/speech/}} for transcribing speech to text. We have tried Kaidi framework as well, but it generally gives worse results. The image features are extracted using CNN of VGG~\cite{vgg16} network while we use a two-layer LSTM with 300 hidden layer nodes each for textual features. Subsequently, image and textual features are subjected to element-wise multiplication and a fully connected layer before the final layer which is a single node regression layer. This layer outputs the score for the speech recognition alternatives.

\begin{figure*}
\vspace{-4mm}
     \begin{center}
     \begin{tabular}{cccc}
     \hspace{-1mm}
     \raisebox{-\totalheight}{\includegraphics[width=0.20\textwidth, height=24mm, trim={0mm 0mm 5mm 0mm},clip]{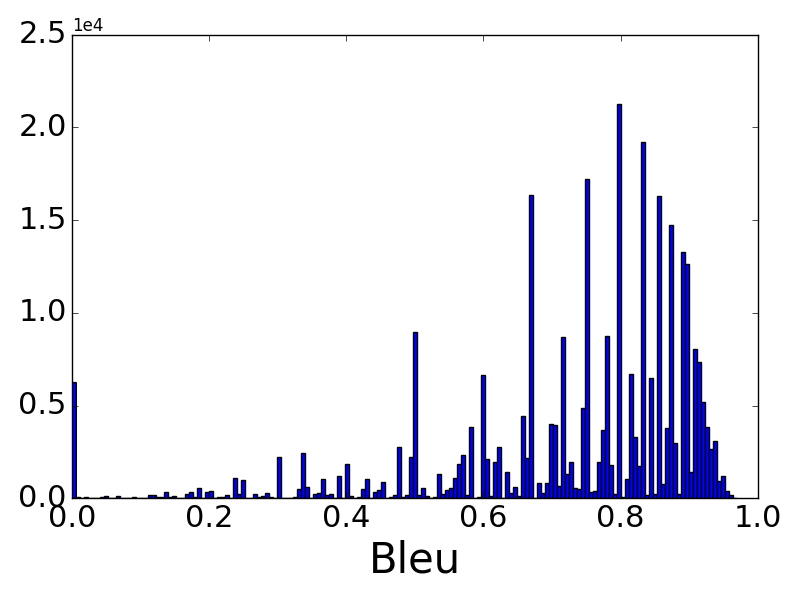}}
      &  \hspace{-2mm}
      \raisebox{-\totalheight}{\includegraphics[width=0.20\textwidth, height=24mm, trim={0mm 0mm 5mm 0mm},clip]{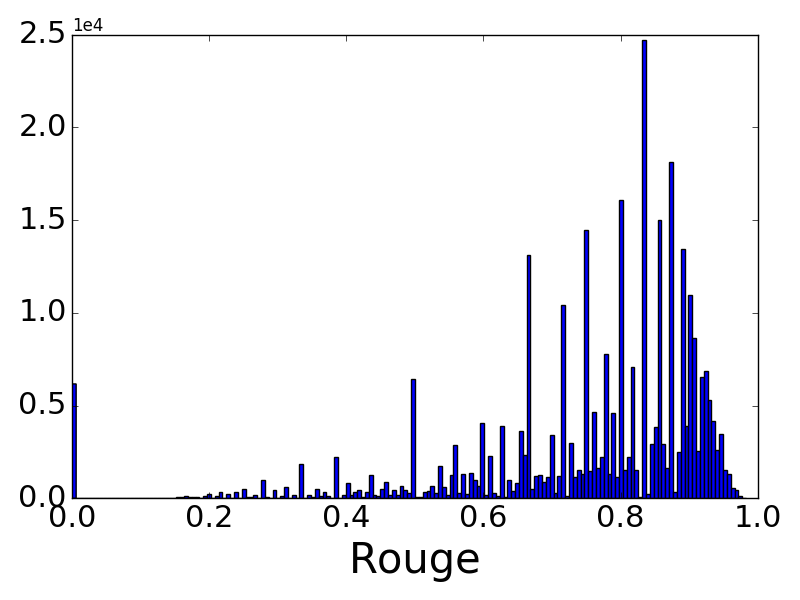}}
      & 
\hspace{-2mm}
       \raisebox{-\totalheight}{\includegraphics[width=0.20\textwidth, height=24mm, trim={0mm 0mm 5mm 0mm},clip]{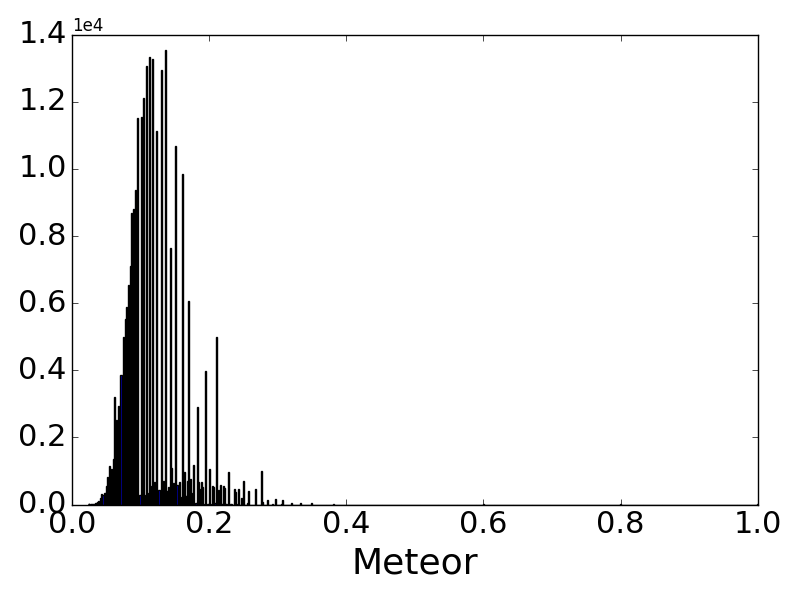}}
        & \hspace{-2mm}
       \raisebox{-\totalheight}{\includegraphics[width=0.20\textwidth, height=24mm, trim={0mm 0mm 5mm 0mm},clip]{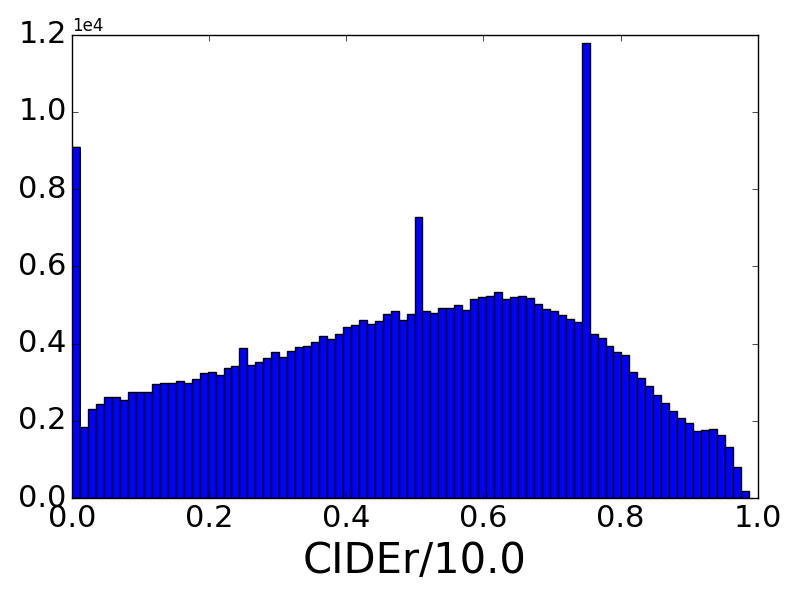}} \\
      \end{tabular}
      \caption{Frequency distributions of scores evaluated by comparing  the text alternatives by ASR for the recorded speech (with 10\% added noise) to ground-truth expressions.}
\label{fig:metrics_hist}
\end{center}
\vspace{-5mm}
\end{figure*}

For training, we use the alternative transcriptions of all the synthetic speech files of GoogleRef train set. For objective scores, we use the following metrics: BLEU~\cite{papineni2002bleu}, ROUGE~\cite{lin2004rouge}, METEOR~\cite{banerjee2005meteor} and CIDEr~\cite{vedantam2015cider}, which are the well-known evaluation metrics in NLP domain. In Figure~\ref{fig:metrics_hist}, we plot the histogram of the scores of the metrics having mapped to the range of $[0,1]$. The google speech alternatives have little differences with the ground truth expressions. We experiment to determine which metric captures these subtle differences. In Figure~\ref{fig:metrics_hist}, the lesser the skewness in the score distribution, the better the metric is sensitive to the subtle differences. CIDEr seems to have a low skew. Consequently, we use CIDEr scores for training the model because they have equi-distribution over the entire range of scores compared to the other three as shown in Figure~\ref{fig:metrics_hist}. In the inference stage, for a given image and the set of alternatives, we use the output regression scores to rank the alternatives. This way, we have the  VGSR.

\subsection{Language-Grounded Object Proposal (LOP)}
\label{subsection:lop}
Given an image and a text expression, the aim of Language Grounded Object Proposal (LOP) is to propose a set of object candidates that  \emph{belong to} the referred class. 

There are numerous object proposal techniques in the literature. For instance, \cite{hu2016natural} uses EdgeBox~\cite{ZitnickECCV14edgeBoxes} for the object proposals;  
\cite{mao2016generation} and \cite{deep:semantic:alignment:cvpr15} use the faster RCNN (FRCNN) object detector \cite{renNIPS15fasterrcnn} and recently Mask-RCNN~\cite{he2017mask} to propose the candidates. However, a direct use of the object proposal methods or object detectors is far from being optimal. General Object Proposal and Detection methods such as EdgeBox and FRCNN are expression agnostic. Because of this, candidates can be proposed from any object class, including those which are totally irrelevant. This leads to an inefficient use of the candidate budget.      

We develop a method for expression-based candidate proposals, with the aim to filter out candidates from irrelevant classes at this stage. We train an LSTM-based approach to associate language expressions to the relevant object class. 
In particular, we train a two-layer LSTM with $300$ hidden layer nodes each. The input of the model is the embedded representation of the expressions by the pre-trained word2vec model~\cite{glove}. The output layer is a classification layer to classify expressions to the corresponding classes of the referred objects. 
The model lets us rank proposals by not only detection scores, but also the relevance scores of their classes to the expressions. 

We first evaluate the accuracy of the class association by our LSTM model. On the GoogleRef dataset, it achieves an accuracy of 90.5\% on the validation set. The high accuracy of the class association implies that LOP is promising.  
Due to this high accuracy, it is reasonable to take the most confident one as the  \emph{relevant} class and the rest as \emph{irrelevant} ones. Given all the detection candidates from the FRCNN detector, we use our LSTM association model to filter out all the detections from the \emph{irrelevant} classes, and only pass those from the \emph{relevant} class to the subsequent component for instance identification. 


\subsection{Language-Grounded Instance Detection (LID)}
\label{subsection:lod}
Given an image, referring expression and a set of bounding box object proposals in the image, our aim of Language Grounded Instance Detection (LID) is to identify the exact instance referred by the given expression.

We employ a generative approach developed by Hu \textit{et al}~\cite{hu2016natural} for this task. Here, the model learns a scoring function that takes features from object candidate regions, their spatial configurations, whole image as global context along with the given referring expression as input and then outputs scores for all candidate regions. The model chooses that candidate which maximizes the generative probability of the expression for the target object. The model architecture is same as that of \cite{hu2016natural}. We extract \textit{fc7} features of VGG-16~\cite{vgg16} net from bounding box location for object features and from whole image for global contextual feature. Textual features are extracted using a embedding layer which is also learned in an end-to-end fashion. For Object Referring with Spoken language (ORSpoken), we use the transcribed text results from VGSR as the input referring expressions for LID (c.f. Figure~\ref{fig:full-pipeline}).

\section{Dataset}
\label{sec:dataset}


Creating large datasets for Object Referring with Spoken Language (ORSpoken) is very labor intensive, given the fact that one need to annotate object location, issue language description and record speech description. A few choices are made in this work to scale the data collection: 1) Following previous work for language-based OR~\cite{mao2016generation,hu2016natural}, we use a simple bounding box to indicate the location of object; and 2) Following the success of synthetic data in many other vision tasks\cite{flow:chair,synthetic:text}, we construct hybrid speech recordings which contain sounding-realistic synthetic speeches for training and real human speeches for evaluation. It is to be noted that real speeches are still recorded in constrained environments as explained below. Two datasets are used in this work, one focusing on scenarios for assistive robots and the other for automated cars. 

\subsection{GoogleRef}
\label{subsec:gref}

\textbf{Object Annotation}. As the first testbed, we choose to enrich the standard object referring dataset GoogleRef~\cite{mao2016generation} with speeches. GoogleRef contains $24,698$ images with $85,474$ referring expressions in the training set and $4,650$ images with $9,536$ referring expressions in the testing set. 

\textbf{Synthetic Speech}. Amazon Polly~\footnote{The voice of Joanna is used: \url{https://aws.amazon.com/polly/}} is used to generate the corresponding speech files for all referring descriptions. This TTS system produces high-quality realistic-sounding speech. It is nevertheless still simpler than real human speech due to the lack of tempo variation and ambient noise. We have thus recorded the real speeches for the testing set of GoogleRef. 
All speech files are in wav format, at 16 kHz.

\begin{table}
\resizebox{1.0\columnwidth}{!}{
\centering
\begin{tabular}{@{}llcccccccc@{}}
\toprule
\multicolumn{2}{c}{Dataset}  & \#Images & \#Objects &Synthetic Speech & Real Speech & On-site Noise \\ \midrule
\multirow{2}{*}{GoogleRef}  & Train & $24698$ & $85474$  &\cmark &  \xmark & \xmark\\
 & Test &$4650$ & $9536$ & \cmark & \cmark & \xmark \\
\multirow{2}{*}{DrivingRef} & Train &$250$ & $750$ & \cmark & \cmark & \cmark \\
 & Test &$250$ & $750$  & \cmark & \cmark & \cmark \\
\bottomrule
\end{tabular}
}
\caption{Statistics of the two datasets.}
\label{tab:dataset-stats}
\vspace{-5mm}
\end{table}


\textbf{Real Speech}. We recorded real speech for all the referring expressions of GoogleRef by crowd-sourcing via Amazon Mechanical Turk(AMT). In AMT web interface, we show five text expressions for each Human Intelligence Task(HIT) and workers are instructed to speak out the expression and record each of the expression separately before submitting the HIT. Workers are asked to speak out as they are talking to a robot. We perform automatic validation by checking the length and volume of the recorded speeches to reject erroneous recordings due to inappropriate use of microphones. After each recording, we display the waveform of the recorded speech and an audio playback tool to hear back the recordings. We also keep a threshold for the amplitude of the audio signal to ensure the recording to attain certain volume level. Our automatic filter only accept the HIT only when all the five recordings are sufficiently long and the amplitude reaches the threshold. To further improve the quality of recordings, we also ran an initial round of recordings to qualify workers; only the qualified workers are allowed to our final recordings. 


\textbf{Noisy Speech}. To further study the performance of the method, we add different levels of noise to the speech files. The noise we mixed with the original speech files is selected randomly from the Urban8K dataset~\cite{urban8k}. This dataset contains 10 categories of ambient noise: air conditioner, car horn, children playing, dog bark, drilling, enging idling, gun shot, jackhammer, siren, and street music. For each original audio file, a random noise file is selected, and combined to produce a noisy audio expression:
\begin{equation}
  E_{noisy} = (1-\beta)*E_{original}  + \beta*Noise
  \label{equ:noise}
\end{equation}
where $\beta$ is the noise level. The noise audio files are sub-sampled to 16 kHz in order to match that of the speech audio file.  Both files are normalized before being combined so that contributions are strictly proportional to the noise level chosen. We study 3 noise levels: 0\%, 10\%, and 30\%.  

\begin{figure}
\centering
\begin{tabular}{cccccc}
\hspace{-1.5mm}
\begin{adjustbox}{valign=t}
  \includegraphics[width=1.0in,height=0.9in]{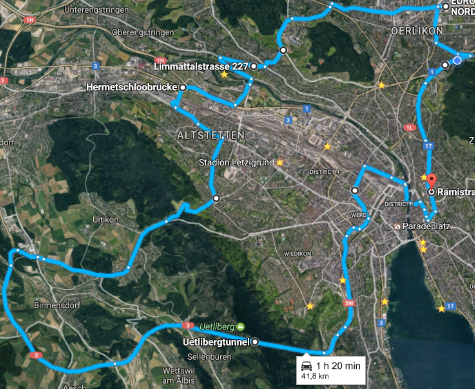}
\end{adjustbox}
&
\hspace{-3mm}
\begin{adjustbox}{valign=t}
\begin{tabular}{@{}c@{}}
  \includegraphics[width=1.0in,height=0.6in]{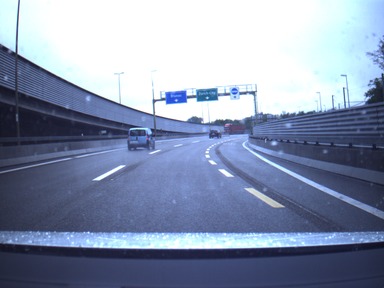}\\[2ex]
  \includegraphics[width=1.0in,height=0.2in, trim={30mm 20mm 20mm 30mm},clip]{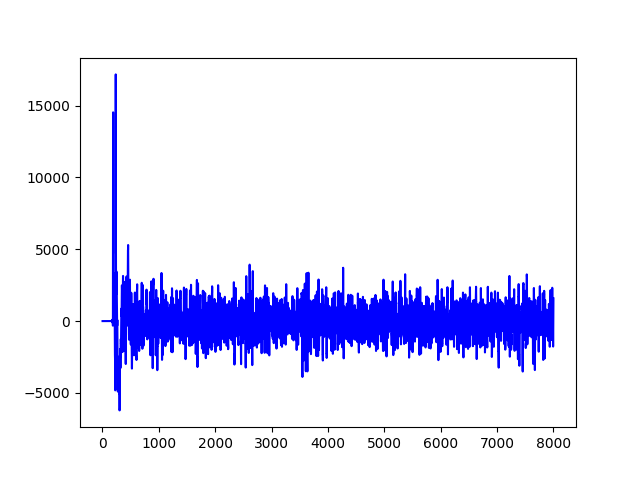}
\end{tabular}
\end{adjustbox}
&
\hspace{-3mm}
\begin{adjustbox}{valign=t}
\begin{tabular}{@{}c@{}}
  \includegraphics[width=1.0in,height=0.6in]{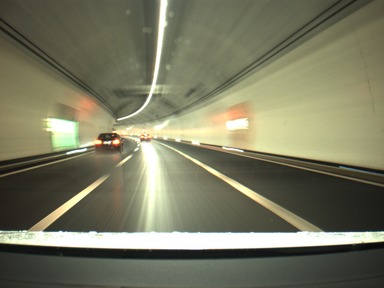}\\[2ex]
  \includegraphics[width=1.0in,height=0.2in, trim={30mm 20mm 20mm 30mm},clip]{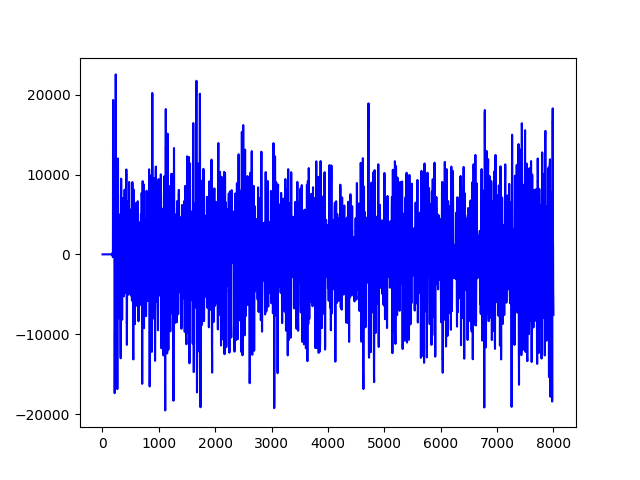}
\end{tabular}
\end{adjustbox}
\\
\text{(a) driving route}  & \text{(b) highway} & \text{(c) tunnel} \\ 
\end{tabular}
\caption{Driving route of our DrivingRef, along with two exemplar visual scenes and the corresponding in-vehicle noise characteristics.}
\label{fig:driving:route}
\vspace{-5mm}
\end{figure}

\subsection{DrivingRef Dataset} 
\label{subsec:drivingdataset}
\textbf{Object Annotation.} In order to have more realistic noise and to evaluate our method in human-vehicle communication, we have created a new dataset DrivingRef for driving scenario. We drove over a 90-minutes route covering highway, residential area, busy city and tunnels. Driving scenes were recorded by a camera mounted on the windshield of the car and the in-vehicle noise was recorded by a microphone mounting on the dashboard of the car.  Figure~\ref{fig:driving:route} shows the driving route, along with samples of the visual scenes and the corresponding in-vehicle noises. We sampled the frames uniformly from the entire length of the recorded video. For annotation, we crowd-sourced to AMT to annotate bounding boxes for the objects on the sampled frames. In the AMT interface, we show the frame and workers are asked to draw tight bounding boxes around the object they would like to refer to. For each bounding box annotated, workers are asked to type a truthful, informative, relevant, and brief expression so that co-observers can find the referred objects easily and unambiguously. For each image, we ask them to annotate four objects along with their referring expressions. In total, we have annotated 1000 objects in 250 diverse images chosen out of the 10,000 recorded images.

\textbf{Synthetic \& Real Speech.} We recorded the text descriptions the same way as described in Section~\ref{subsec:gref}.

\textbf{Noisy Speech} In-vehicle noise characteristics are ascribed to many factors such as driving environment (i.e. highway, tunnel), driving states (i.e. speed, acceleration), and in-vehicle noise (i.e. background music). During the course of data collection for DrivingRef, we also recorded the in-vehicle noise from inside the vehicle. We added this recorded real in-vehicle noise to the speeches according to Eq.~\ref{equ:noise}. Timestamps of images and the recorded noise audio are used to obtain the in-place noise clips for the corresponding images (visual scenes). 
In total, we have qualified 23 workers for the recording task. The statistics of the two datasets can be found in Table~\ref{tab:dataset-stats}.

\begin{table}[!tb]
\resizebox{1.0\columnwidth}{!}{
\centering
\begin{tabular}{@{}cclcccccc@{}}
\toprule 
 & Noise & Method  & METEOR $\uparrow$ & ROUGE $\uparrow$ & CIDEr $\uparrow$ &  BLEU1 $\uparrow$     \\ \midrule
 \parbox[t]{2mm}{\multirow{6}{*}{\rotatebox[origin=c]{90}{Synthetic}}} & \multirow{ 2}{*}{0\%}  
& Google API& 0.570 & 0.880 & 7.416 &0.890  \\
& & VGSR & \textbf{0.591}& \textbf{0.906} &\textbf{7.806} &\textbf{0.910}   \\
\cdashline{2-7}
&   \multirow{ 2}{*}{10\%}  & Google API & 0.536& 0.838 &6.901 &0.846   \\
& & VGSR& \textbf{0.544}& \textbf{0.854} &\textbf{7.048} &\textbf{0.861}  \\
\cdashline{2-7}
&   \multirow{ 2}{*}{30\%}  & Google API& 0.437& 0.683 &5.410 &0.679   \\
& & VGSR& \textbf{0.455}& \textbf{0.714} &\textbf{5.720} &\textbf{0.709}  \\
\cline{2-7}

\parbox[t]{2mm}{\multirow{6}{*}{\rotatebox[origin=c]{90}{Real}}}& \multirow{ 2}{*}{0\%}  
& Google API & 0.391 &0.661 &4.700 &0.662  \\
& & VGSR &\textbf{0.409}& \textbf{0.693} &\textbf{5.058} &\textbf{0.689} \\
\cdashline{2-7}
& \multirow{ 2}{*}{10\%}  & Google API & 0.364& 0.619&4.255 &0.618  \\
& & VGSR & \textbf{0.386}& \textbf{0.654} &\textbf{4.689} &\textbf{0.651} \\
\cdashline{2-7}
& \multirow{ 2}{*}{30\%}  & Google API & 0.286& 0.486&3.162 &0.464   \\
& & VGSR&\textbf{0.305}&\textbf{0.521} &\textbf{3.526}&\textbf{0.495} \\
\bottomrule
\end{tabular}
}
\caption{Comparison of speech recognition results with ground truth text using different standard evaluation metrics, evaluated on GoogleRef dataset.}
\label{tab:text-eval-gref}
\end{table}

\begin{table}
\resizebox{1.0\columnwidth}{!}{
\centering
\begin{tabular}{@{}cclcccccc@{}}
\toprule
& Noise & Method  & METEOR $\uparrow$ & ROUGE $\uparrow$ & CIDEr $\uparrow$ &  BLEU1 $\uparrow$     \\ \midrule
 \parbox[t]{2mm}{\multirow{6}{*}{\rotatebox[origin=c]{90}{Synthetic}}} & \multirow{ 2}{*}{0\%}  
& Google API & 0.548 & 0.869 & 6.922 & 0.868   \\
& & VGSR &\textbf{0.577}& \textbf{0.899} &\textbf{7.735} & \textbf{0.899}   \\
\cdashline{2-7}
 & \multirow{ 2}{*}{10\%}  & Google API& 0.534& 0.855 &6.63 & 0.857   \\
& & VGSR & \textbf{0.542}& \textbf{0.868} &\textbf{7.114} &\textbf{ 0.871}    \\
\cdashline{2-7}
& \multirow{ 2}{*}{30\%} & Google API & 0.499& 0.816 & 6.149 & 0.820   \\
& & VGSR & \textbf{0.517}& \textbf{0.837} &\textbf{6.703} & \textbf{0.837}  \\
\cline{2-7}

\parbox[t]{2mm}{\multirow{6}{*}{\rotatebox[origin=c]{90}{Real}}}& \multirow{ 2}{*}{0\%}  
& Google API  & 0.372 & 0.688 & 4.331 & 0.674 \\
& & VGSR & \textbf{0.392} & \textbf{0.716} & \textbf{4.746} & \textbf{0.697}\\
\cdashline{2-7}& \multirow{ 2}{*}{10\%}  & Google API & 0.360 & 0.670 & 4.175 & 0.655   \\
& & VGSR &\textbf{0.380}&\textbf{0.705} &\textbf{4.648}&\textbf{0.682} \\
\cdashline{2-7}
& \multirow{ 2}{*}{30\%}  & Google API & 0.290 & 0.549 & 3.159 & 0.516   \\
& & VGSR &\textbf{0.303}&\textbf{0.574} &\textbf{3.449}&\textbf{0.536} \\
\bottomrule
\end{tabular}
}
\caption{Comparison of speech recognition results with ground truth text using different standard evaluation metrics, evaluated on DrivingRef dataset.}
\label{tab:text-eval-trace}
\end{table}

\begin{figure*}
\centering
\begin{tabular}{cc}
\noindent
\includegraphics[width=0.38\textwidth,trim={.05\textwidth} {.00\textwidth} {0.05\textwidth} {.05\textwidth},clip]{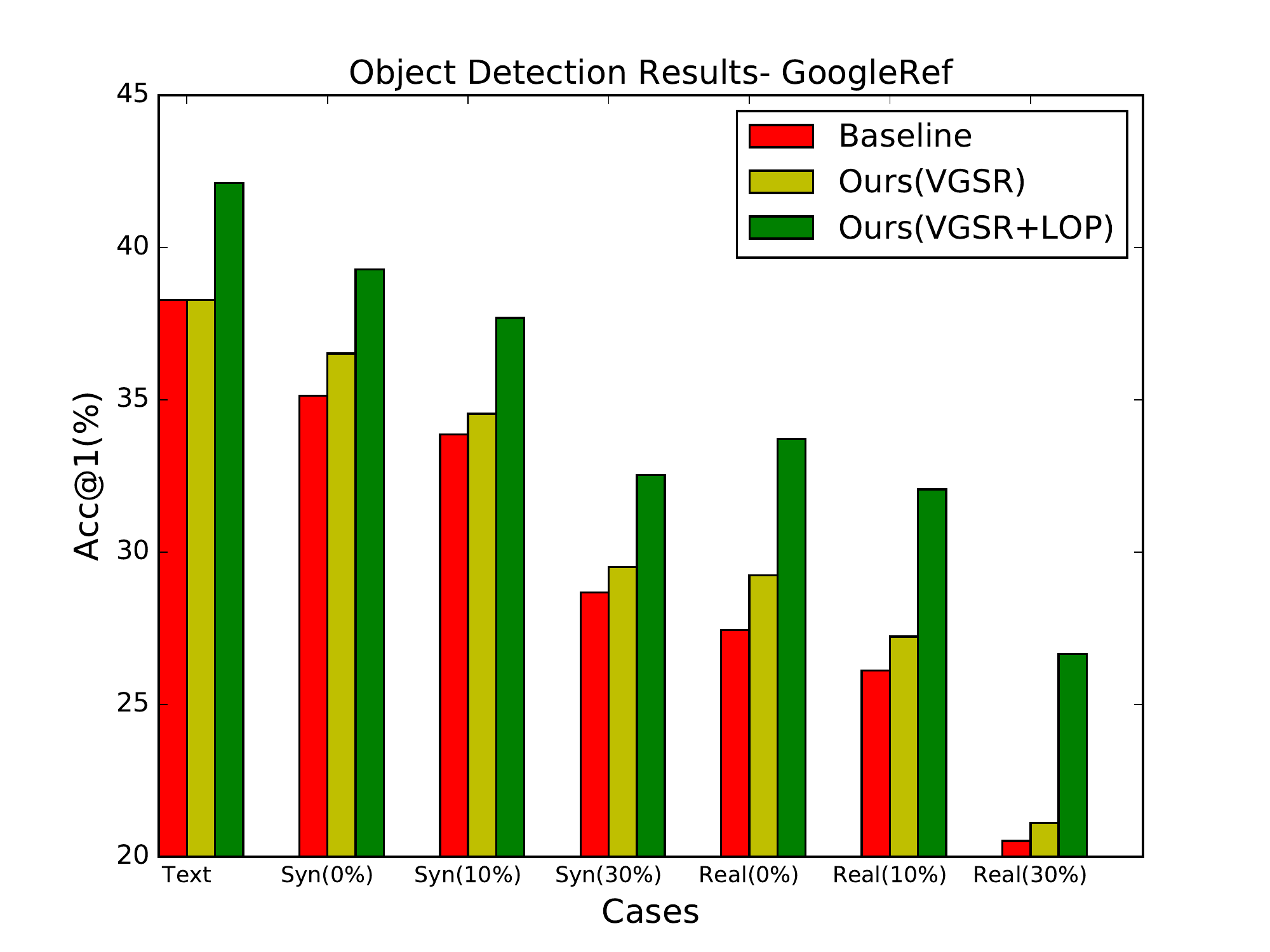} &
\noindent 
\includegraphics[width=0.38\textwidth,trim={.05\textwidth} {.00\textwidth} {0.05\textwidth} {.05\textwidth},clip]{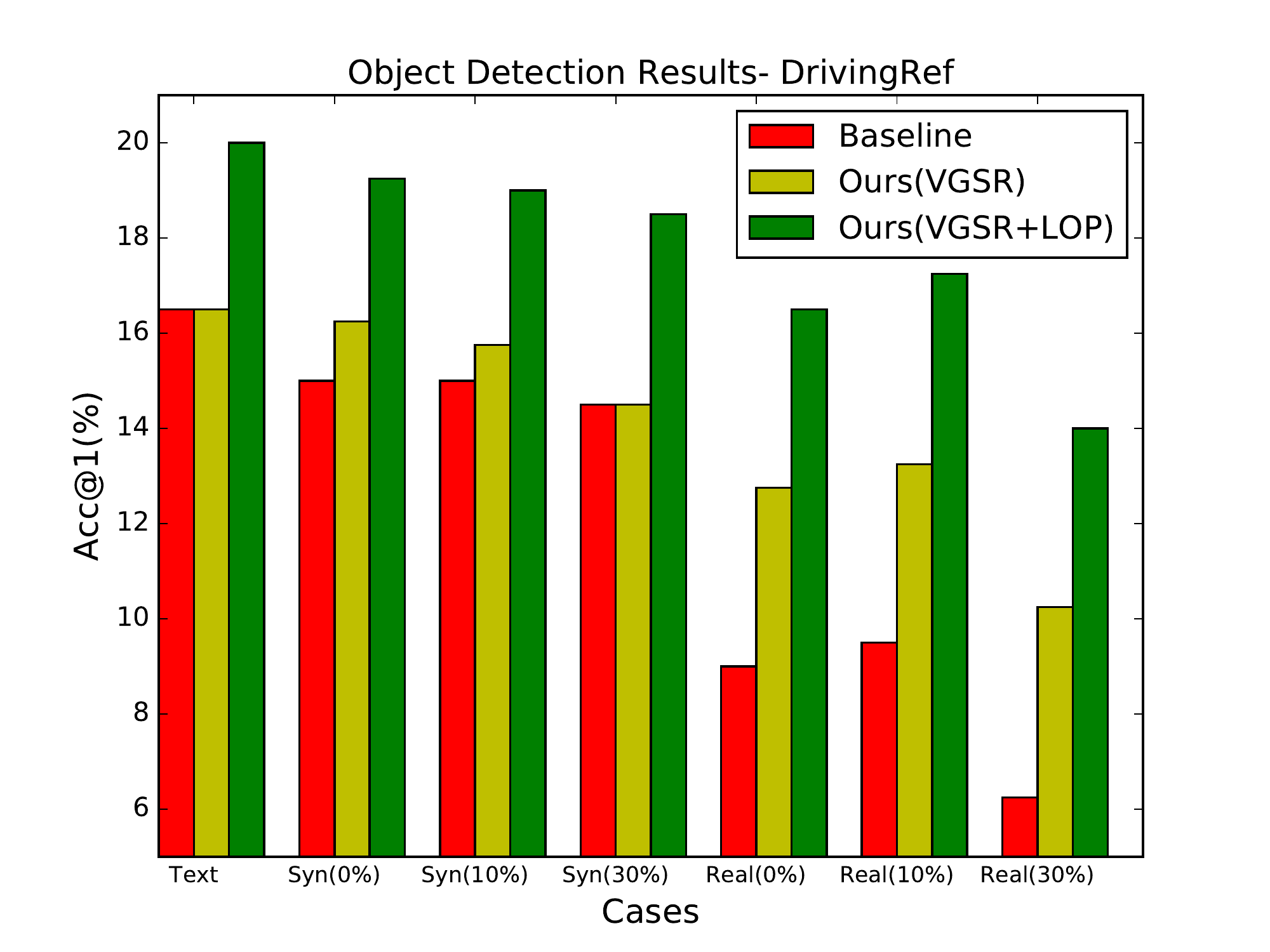}\\

\end{tabular}  \vspace{-2mm}
\caption{Comparison plot of Ours(VGSR) and Ours(VGSR+LOP) with the baseline method on a) GoogleRef b) DrivingRef. We show the accuracy of object recognition for LID task from (i) text, (ii) transcribed text results from Synthetic and Real Speech at different noise levels. Over the baseline method, we introduce the contribution of VGSR for Ours(VGSR) and later include LOP for Ours(VGSR+LOP).}
\label{fig:googleref_barplot}
\vspace{-5mm}
\end{figure*}

\section{Experiments}
\label{sec:experiment}
We first evaluate the overall performance of our approach, and then conduct detailed evaluation for the two components Visually Grounded Speech Recognition (VGSR) and Language Grounded Object Proposal (LOP). 



\subsection{Evaluation Metric}
For the final performance of our method for Object Referring with Spoken Language (ORSpoken), we use $Acc@1$ by following~\cite{hu2016natural}. $Acc@1$ refers to the percentage of top scoring candidate(rank-1) being a true detection. A candidate is considered as a true detection if IoU computed between the predicted bounding box and the ground truth box is more than 0.5. We evaluate the performance of VGSR based on standard criterions such as METEOR~\cite{banerjee2005meteor}, ROUGE~\cite{lin2004rouge}, BLEU~\cite{papineni2002bleu} and CIDEr~\cite{vedantam2015cider}. We evaluate the results of VGSR in comparison to the ground truth referring expressions generated by humans using which we record/generate the speech files. For LOP, we use Intersection over Union (IoU) and recall to evaluate the performance. 

\begin{figure*}[!tb]
\vspace{-5mm}
     \begin{center}
     \begin{tabular}{cccc}
     \raisebox{-\totalheight}{\includegraphics[width=0.24\textwidth, height=20mm, trim={30mm 40mm 60mm 20mm},clip]{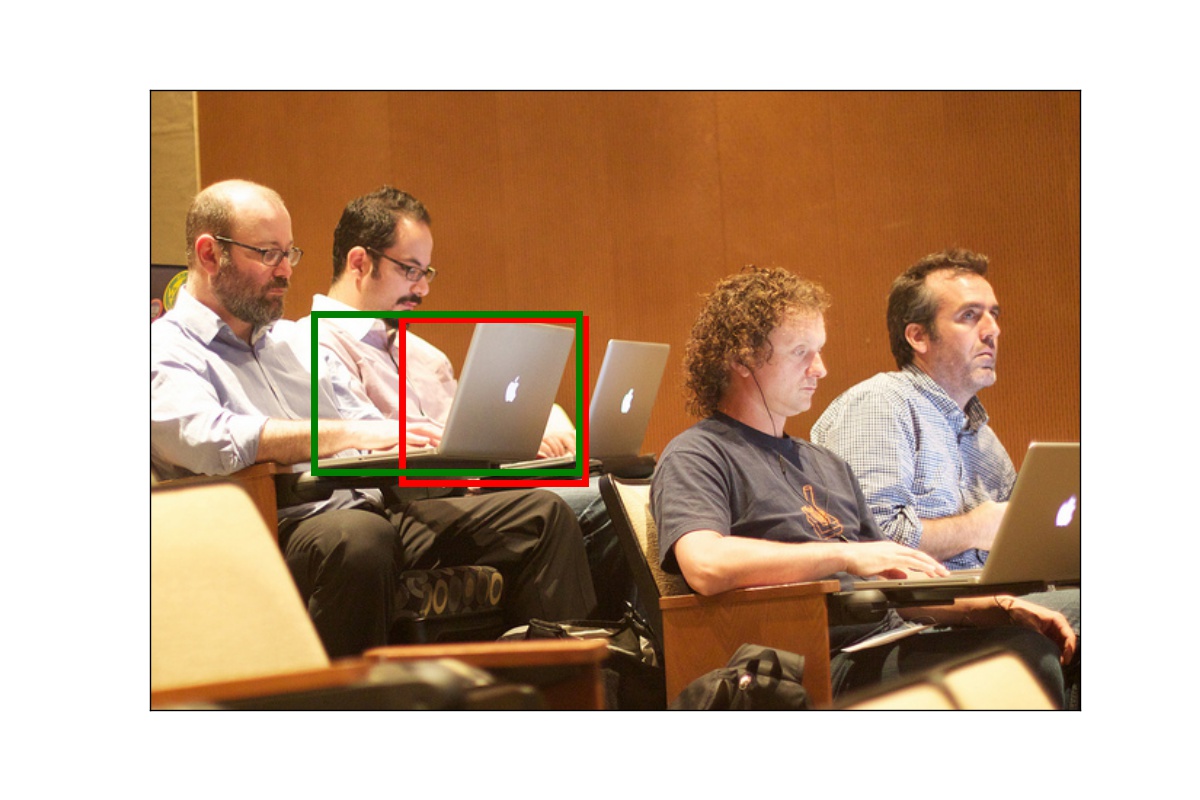}}
      & \hspace{-2mm}
      \raisebox{-\totalheight}{\includegraphics[width=0.24\textwidth, height=20mm, trim={50mm 30mm 50mm 20mm},clip]{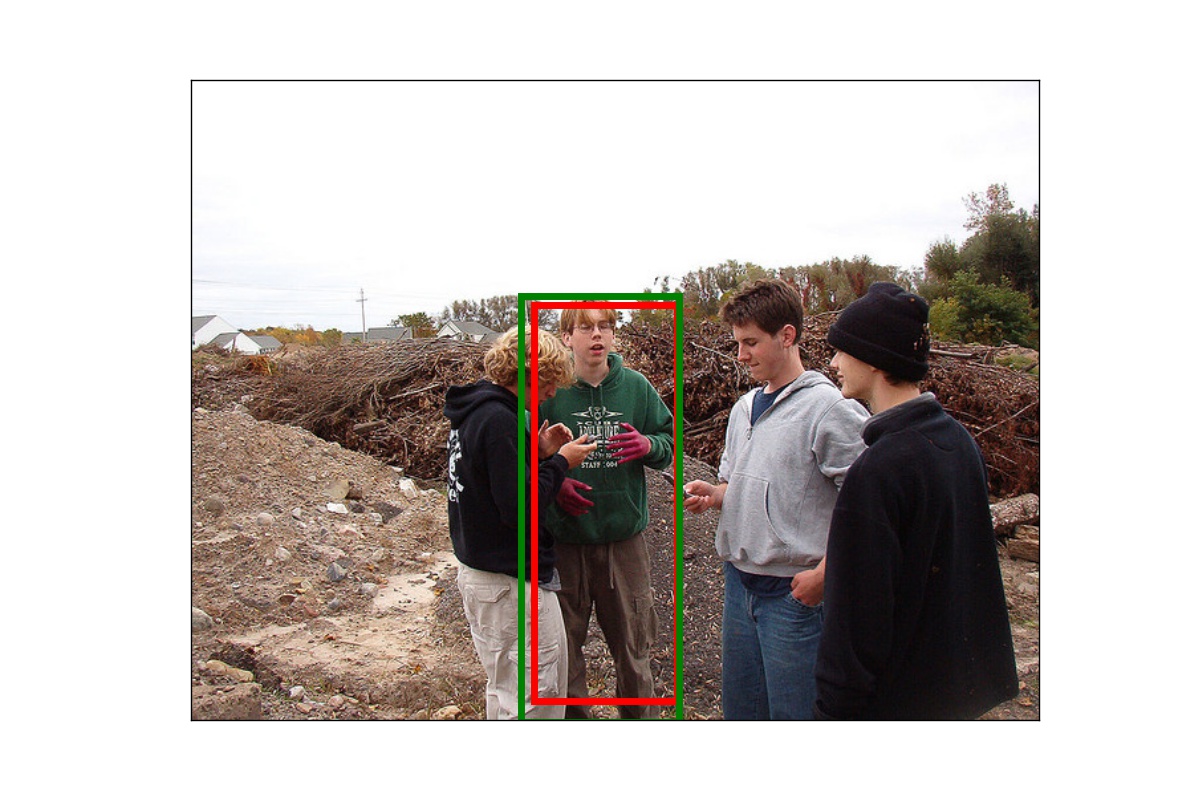}}& \hspace{-2mm}
       \raisebox{-\totalheight}{\includegraphics[width=0.24\textwidth, height=20mm, trim={50mm 50mm 60mm 20mm},clip]{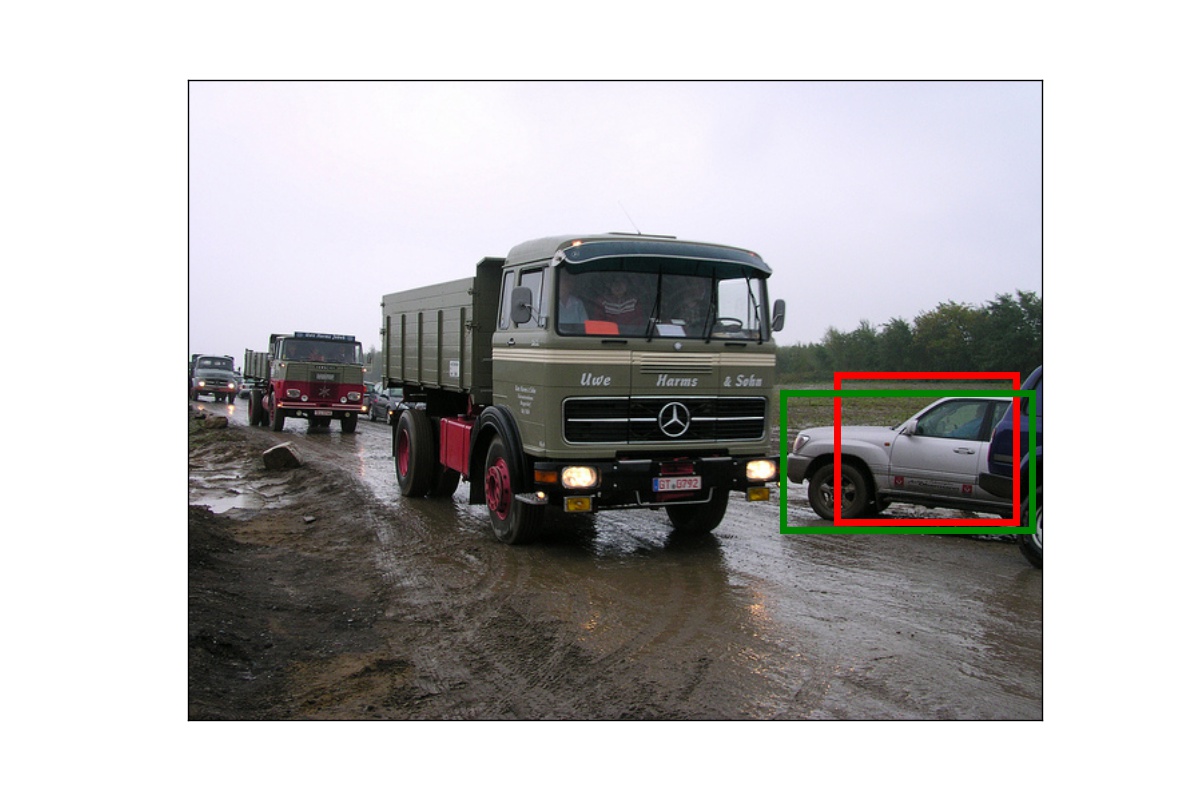}} & \hspace{-2mm}
        \raisebox{-\totalheight}{\includegraphics[width=0.24\textwidth, height=20mm, trim={50mm 20mm 20mm 20mm},clip]{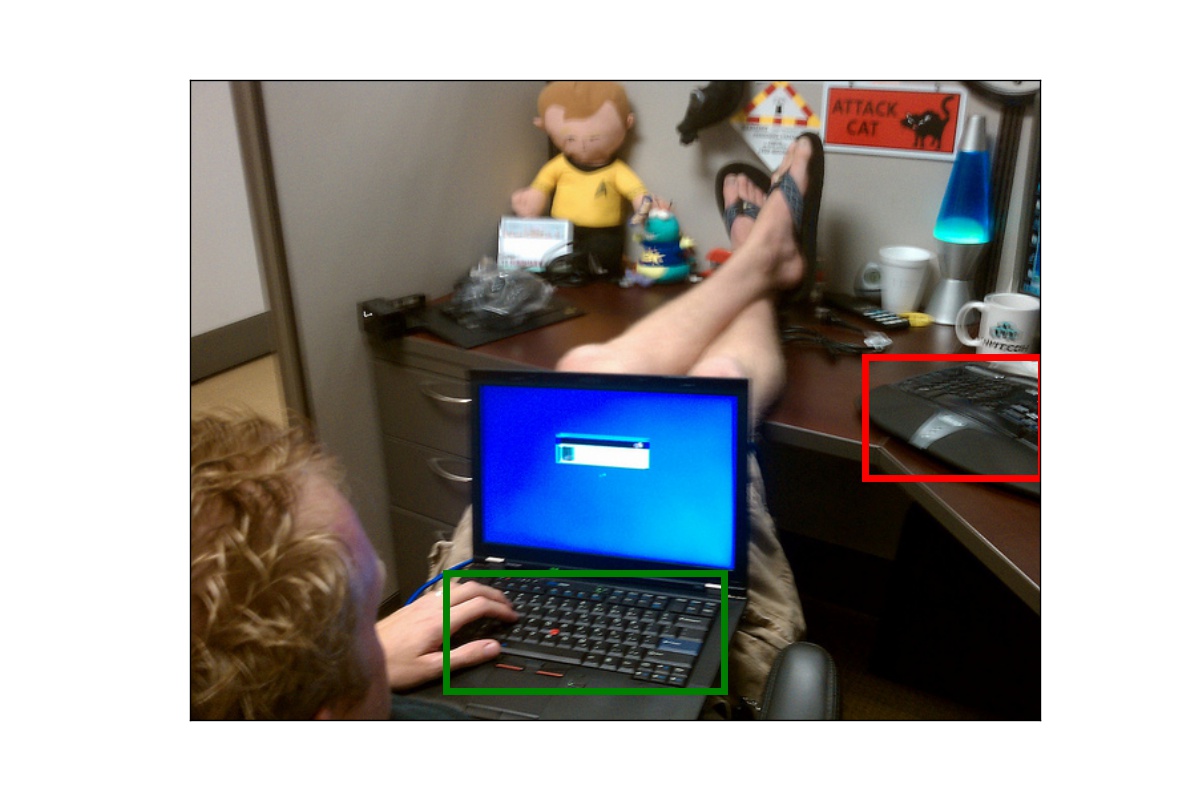}} \\ 
        \hspace{-1mm}    \tiny{\textcolor{red}{Speech:}laptop the bed man is using} &  \hspace{-2mm}\tiny{\textcolor{red}{Speech:}green t-shirt talking to two other men} &  \hspace{-2mm} \tiny{\textcolor{red}{Speech:}the silver vehicle next to a green truck} &  \hspace{-2mm} \tiny{\textcolor{red}{Speech:}keyboard on the laptop}\vspace{-2mm} \\ 
        \tiny{\textcolor{green}{GT:}laptop the bald man is using} &  \hspace{-2mm}\tiny{\textcolor{green}{GT:}a man in green t shirt talking to two other men} &  \hspace{-2mm} \tiny{\textcolor{green}{GT:}a silver vehicle next to a green truck} &  \hspace{-2mm} \tiny{\textcolor{green}{GT:}keyboard on the laptop}\\ 

 \hspace{-1mm}
     \raisebox{-\totalheight}{\includegraphics[width=0.24\textwidth, height=20mm, trim={30mm 40mm 60mm 20mm},clip]{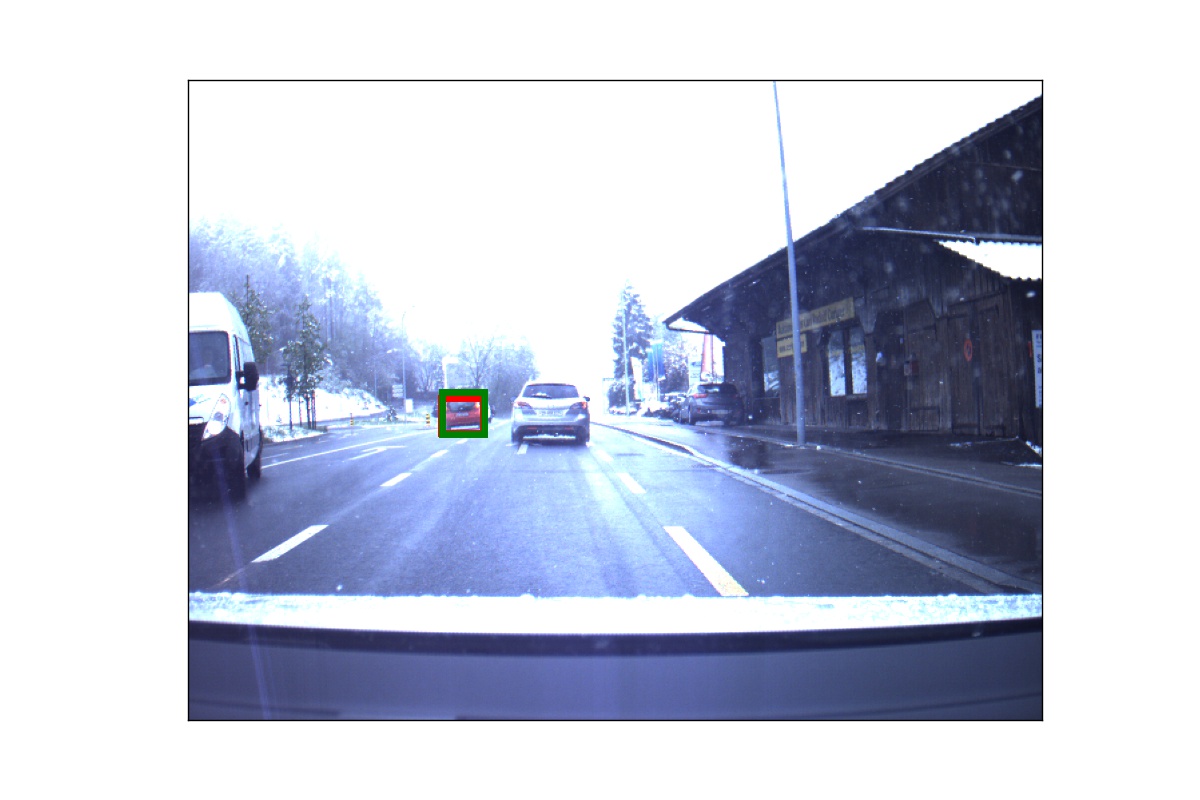}}
      &     \hspace{-2mm}
      \raisebox{-\totalheight}{\includegraphics[width=0.24\textwidth, height=20mm, trim={50mm 30mm 50mm 20mm},clip]{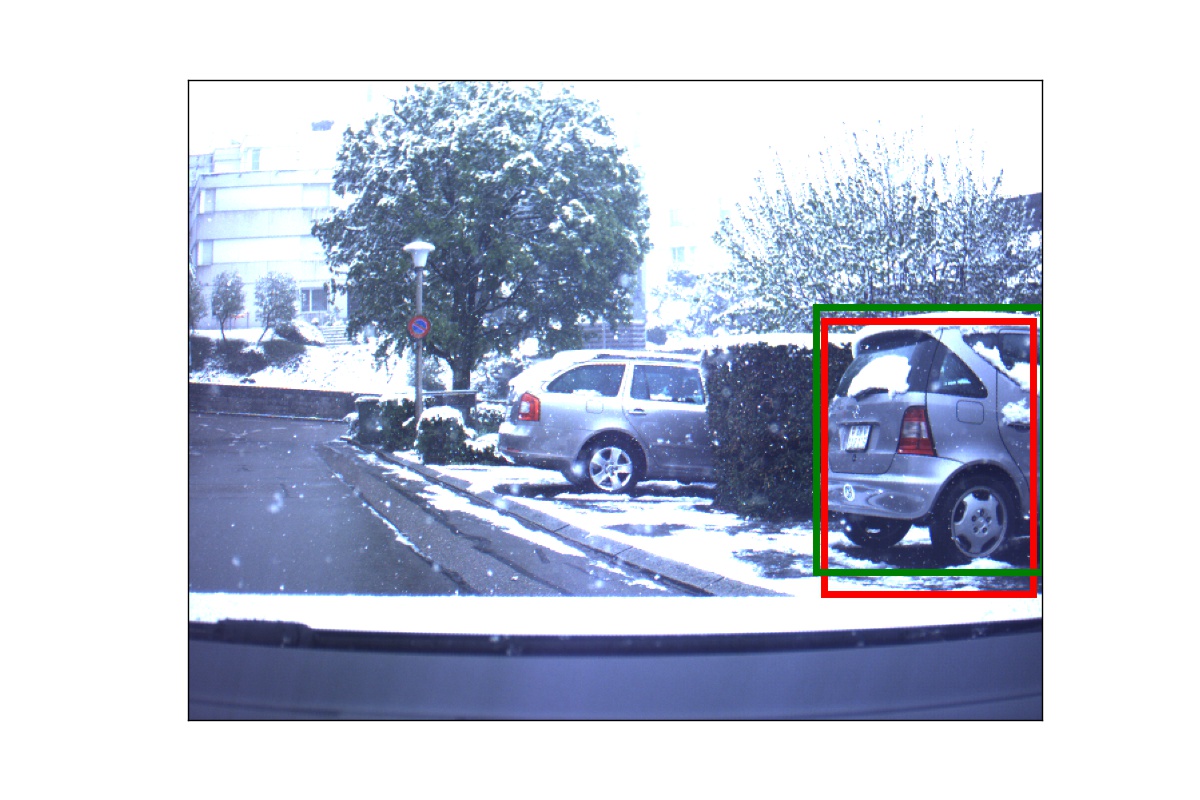}}
      &  \hspace{-2mm}
       \raisebox{-\totalheight}{\includegraphics[width=0.24\textwidth, height=20mm, trim={50mm 50mm 60mm 20mm},clip]{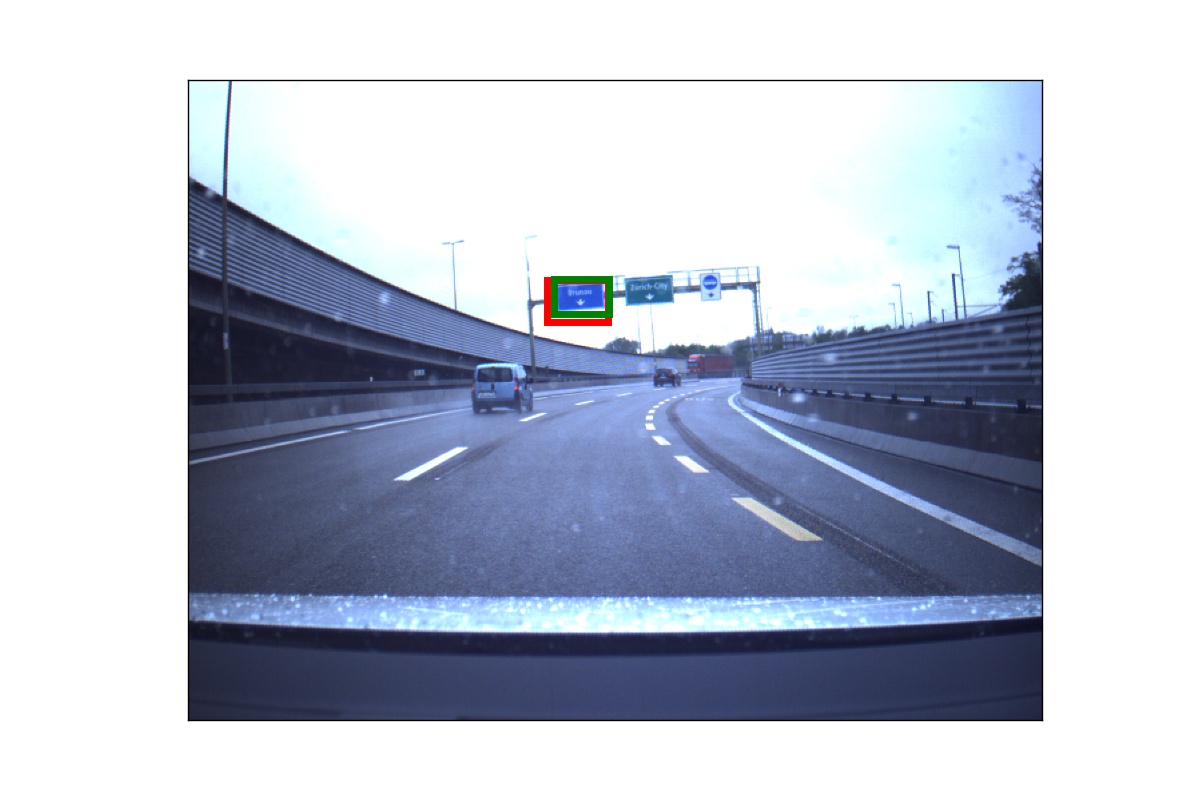}}
         &  \hspace{-2mm}
        \raisebox{-\totalheight}{\includegraphics[width=0.24\textwidth, height=20mm, trim={50mm 20mm 20mm 20mm},clip]{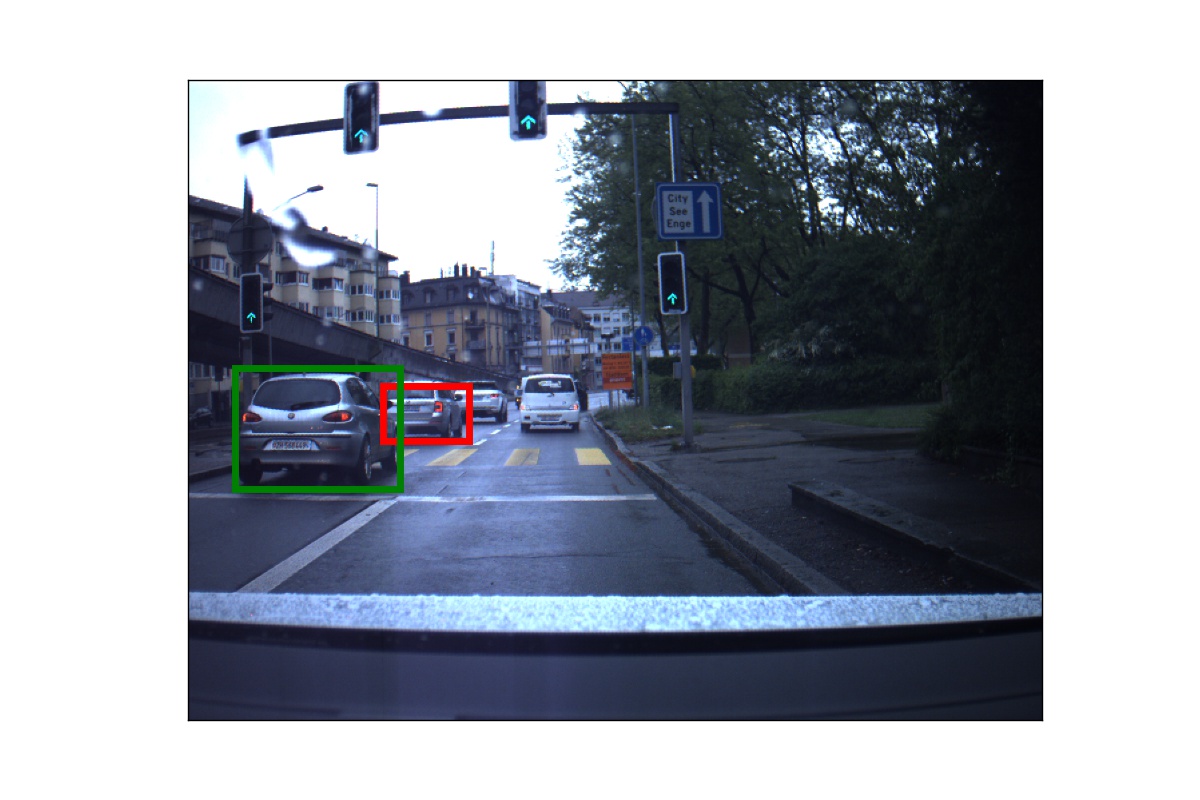}} \\ 
        \hspace{-1mm}    \tiny{\textcolor{red}{Speech:}red car moves ahead on} &  \hspace{-2mm}\tiny{\textcolor{red}{Speech:}silver car parked close to us} &  \hspace{-2mm} \tiny{\textcolor{red}{Speech:}blue signboard on the top left most} &  \hspace{-2mm} \tiny{\textcolor{red}{Speech:}lane} \vspace{-2mm} \\ 
        \tiny{\textcolor{green}{GT:}red car moves ahead on left} &  \hspace{-2mm}\tiny{\textcolor{green}{GT:}silver car parked close to us on our right} &  \hspace{-2mm} \tiny{\textcolor{green}{GT:}blue signboard on the top left most side} &  \hspace{-2mm} \tiny{\textcolor{green}{GT:}gray car just moves past the traffic signal on the left lane}\\  
      \end{tabular}  
      \caption{Qualitative results on GoogleRef (top) and DrivingRef (bottom)) with real speech of the text descriptions and the corresponding prediction box shown in \textcolor{red}{Red} and ground truth text and the corresponding bounding box are shown in \textcolor{green}{Green}.}
\label{fig:qual_gref}
\end{center}
\end{figure*}

\subsection{Object Referring with Spoken Language}
Since there is no previous work to compare to, we compare with a baseline method. For the \textbf{baseline model}, we use GoogleAPI to transcribe the speech data to text. We then use the object referring model of \cite{hu2016natural} to select the object candidates proposed by FRCNN detector.  Coming to \textbf{Our approach} as described in Section~\ref{sec:approach}, we use VGSR model to transcribe the speech data to text. Following that, we use the same object referring model of \cite{hu2016natural} but we propose the object candidates using our method LOP (as in Figure~\ref{fig:obj_prop_comp}). 
We also use text-based object referring as our references for both the baseline method and our method. That is, clean text is given as input to the two approaches without employing speech recognition. The evaluation is conducted for both synthetic speech and real speech, and at different noise levels. 

We illustrate the results of ORSpoken in Figure~\ref{fig:googleref_barplot}a for GoogleRef dataset and in Figure~\ref{fig:googleref_barplot}b for DrivingRef dataset.
We can observe that detection results from the text remains as an upper bound for the object localization accuracy since all other cases use the transcribed texts from either synthetic or real speech data of the same text. From Figure.~\ref{fig:googleref_barplot}, we observe that Ours(VGSR) and Ours(VGSR+LOP) clearly outperforms the baseline model in all cases in both GoogleRef and DrivingRef dataset. Taking the case of Ours(VGSR+LOP), it performs better than the baseline model in $Acc@1$ in all the cases of synthetic and real speech with 0\%, 10\% and 30\% noise. For instance, Case 1: $Acc@1$ improves by 1.526\% for 10 proposals from baseline to Ours  for the synthetic speech without 30\% noise in Figure~\ref{fig:googleref_barplot}a, Case 2: $Acc@1$ improves by 3.925\% for 10 proposals from baseline to Ours for real speech data with 30\% noise. Similar observation can be seen for DrivingRef dataset in Figure~\ref{fig:googleref_barplot}b also. From Figure.~\ref{fig:googleref_barplot}, we can infer that each component of \emph{Our approach}(VGSR\&LOP) contributes to the improvement of overall performance of ORSpoken task. This shows that \emph{Our approach} in realistic scenarios helps accommodate for a better object localization than in the synthetic cases which shows the suitability of our method to the real world. 

We have added some qualitative results of \emph{our approach} in GoogleRef and DrivingRef dataset in Figure~\ref{fig:qual_gref}. We have the object detection results represented as bounding boxes along with the corresponding transcribed text and ground truth texts for GoogleRef and DrivingRef dataset. We have also added some failure cases in the last column. We see that object detection model(LID) fails for GoogleRef as the predicted box is on a different `\emph{keyboard}' while in the second row, we observe that speech recognition itself fails for an appropriate transcription.

Apart from our pipelined approach as described in Section~\ref{sec:approach}, we also made an attempt towards an end-to-end solution to the task of ORSpoken. We tried to learn directly from from raw speech and image, and avoiding the intermediate stage of speech recognition, similar to the `end-to-end' approach proposed in ~\cite{speechVQA} for speech-based visual question answering. We extract the same set of visual features as in the case of our pipelined approach. 
For the speech part, we pass the raw speech signal through a series of 1D Convolutional Neural networks and a LSTM layer that outputs the vector representation for speech part. We observe that this straightforward solution does not performs better than our pipelined method. This is due to the lack of a large dataset to learn an end-to-end model for the complex task ORSpoken. Our pipelined approach has less complex sub-tasks (VGSR\&LOP) and is able to  leverage the available resources for speech recognition and object recognition.

\subsection{Visually-grounded Speech Recognition(VGSR)}
We conducted experiments of VGSR on GoogleRef and DrivingRef dataset. GoogleRef dataset comprises of $85,474$ referring expressions from $24,698$ images of MSCOCO dataset for training and $9,536$ expressions from $4,650$ images for testing. DrivingRef dataset comprises of $1500$ referring expressions from $500$ images, which we split 50-50 for training and testing. As described in Section~\ref{sec:dataset}, we generate synthetic and real speech data from referring expressions of GoogleRef dataset and add 10\% and 30\% noise content to speech data for various experiments. These speech data is transcribed back to text expressions using Google API. For each speech data, we receive 5 different alternatives from the API~\footnote{In some cases, the number of alternatives are less than 5. We find that this happens when the speech recognition engine is very confident in the answers provided.}. We evaluate these transcribed text results of Speech Recognition (SR) using standard metrics including METEOR, ROUGE, CIDEr,  BLEU1, as shown in Table~\ref{tab:text-eval-gref}. Google API in Table~\ref{tab:text-eval-gref} represents the results of the rank-1 transcribed text among the alternatives obtained. Comparing the rank-1 SR results from Google API under multiple noise levels, we notice that the performance of SR decreases with the increase of noise levels, as in Table~\ref{tab:text-eval-gref}. We also see that better performance is obtained for SR on synthetic speech data compared to real speech data. This is due to the complexity of real speech data contributed by ambient noise, tempo variations, emotions, and accent.   

\begin{figure}[!tb]
  \centering
  \includegraphics[width=0.4\textwidth,trim={.05\textwidth} {.00\textwidth} {0.05\textwidth} {.05\textwidth},clip]{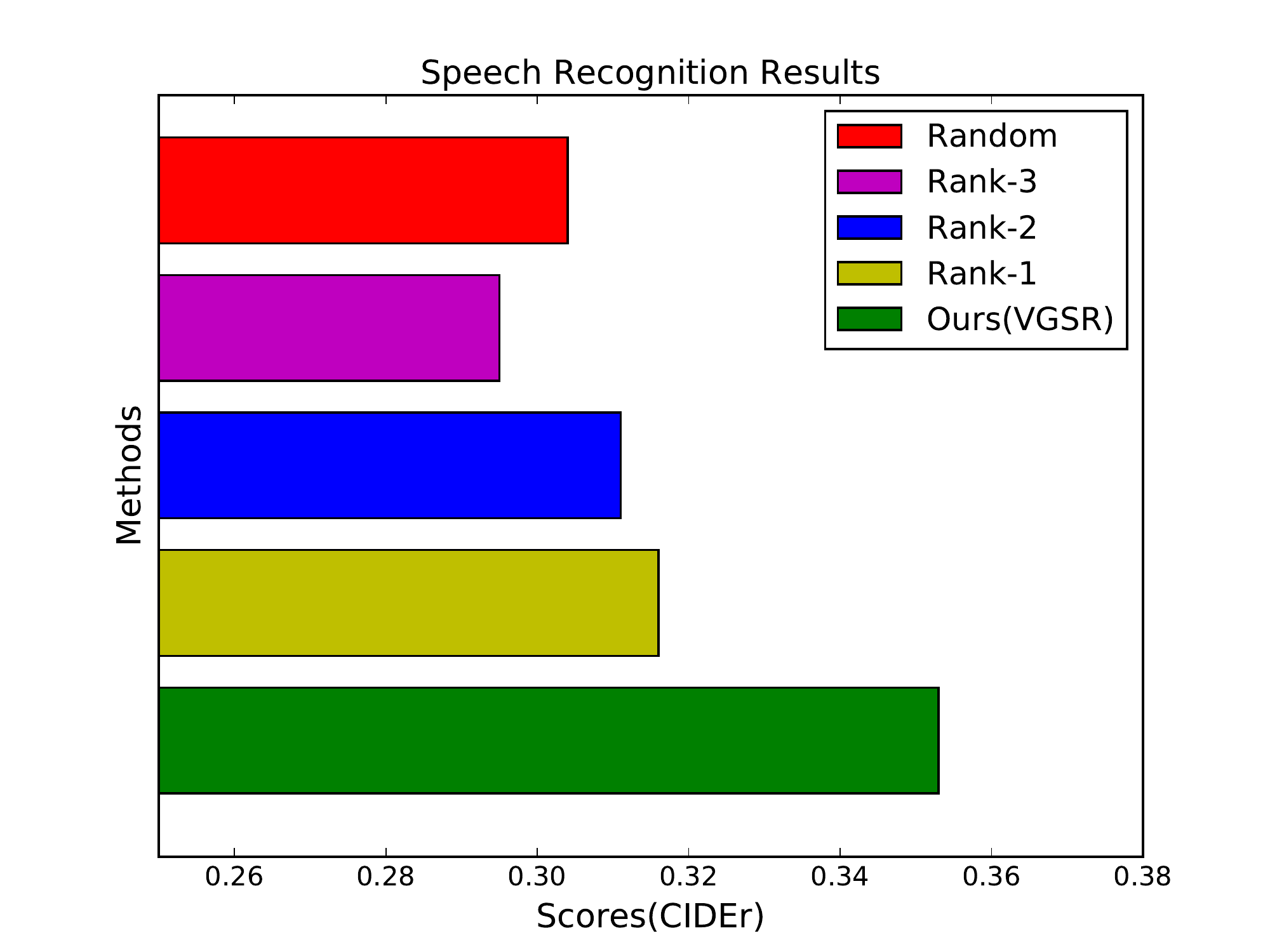}
    \caption{Comparison of Ours(VGSR) to other baselines on GoogleRef dataset; random selection (Random), rank-1, rank-2 and rank-3 results provided by Google SR API are considered. CIDEr is used for the evaluation, for which larger means better. }
  \label{fig:barplot}
\end{figure}

Coming to VGSR, we re-rank the set of text alternatives obtained from SR API, conditioned on the visual context. In Figure~\ref{fig:barplot}, we have compared the SR performances of rank-1, rank-2 and rank-3 transcribed text results from Google API along with performance of VGSR. This experiment is conducted on GoogleRef real speech data with 30\% noise level. We observe that \textit{Ours}:VGSR outperforms all the rank based selections from API. For detailed analysis, we conducted experiments of VGSR over synthetic and real speech data and over multiple noise levels as shown in Table~\ref{tab:text-eval-gref} and Table~\ref{tab:text-eval-trace}. We observe that VGSR performs significantly better in all the cases. For the training of VGSR regression network, we use ground truth labels calculated from CIDEr metric as described in Section~\ref{subsection:vgsr}. We have conducted experiments on GoogleRef dataset in Table~\ref{tab:text-eval-gref} and DrivingRef dataset in Table~\ref{tab:text-eval-trace} which too provide similar observations of superior performance of VGSR. Thus, our experiments show that VGSR performs better than the performance of SR without the context of visual input. We believe that in future complex models can be learned to jointly optimize speech and image context to get an even better performance.

\subsection{Language-grounded Object Proposal (LOP)}

The task of LOP is to propose a set of bounding box locations of the object candidates in the image, given the image and a referring text expression of an object. 

As described in Section~\ref{subsection:lop}, given the text description and the image, LOP uses FRCNN detections to create the first set of proposals. FRCNN is trained for object detection task on MSCOCO dataset for the GoogleRef dataset while we use Cityscapes dataset for the DrivingRef dataset. These choices are due to the fact that object classes are roughly shared between MSCOCO and GoogleRef, and between CityScapes and DrivingRef. Later, LOP filters them to yield candidates of the \emph{relevant} class predicted by the specifically trained LSTM model producing class specific object proposals.

\begin{figure}
\centering
\begin{tabular}{ccc}
\noindent
\includegraphics[width=0.5\linewidth]{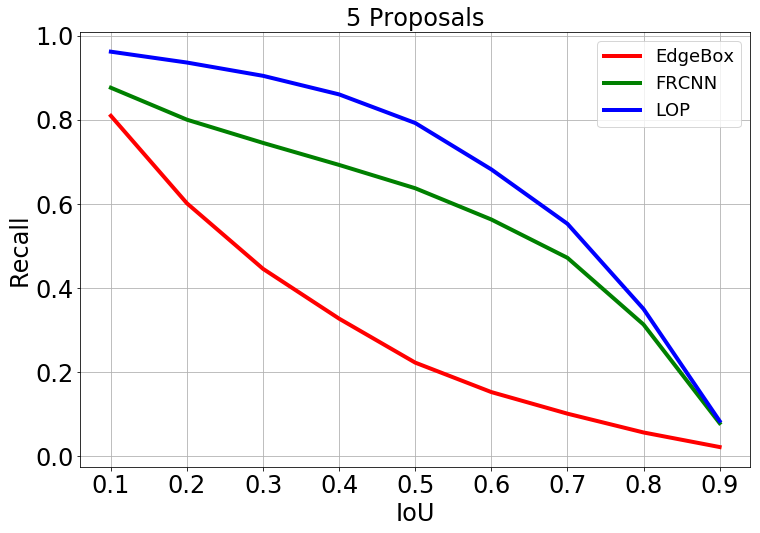}
\noindent 
\includegraphics[width=0.5\linewidth]{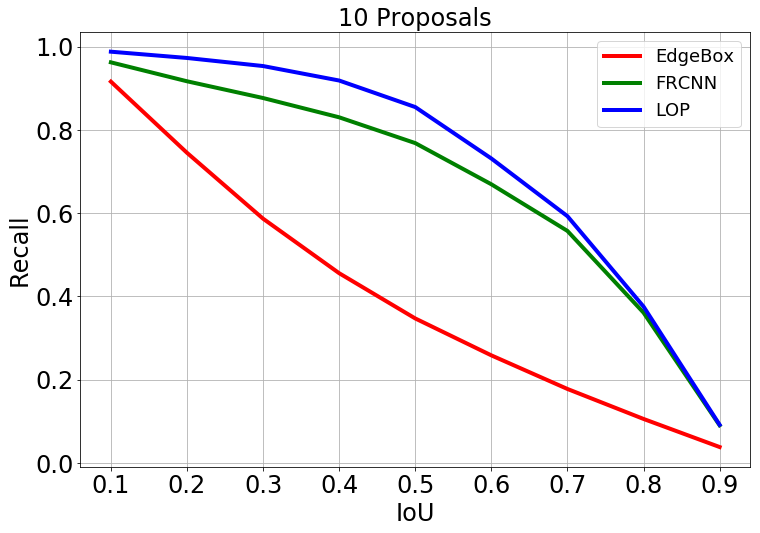}\\
\noindent 
\includegraphics[width=0.5\linewidth]{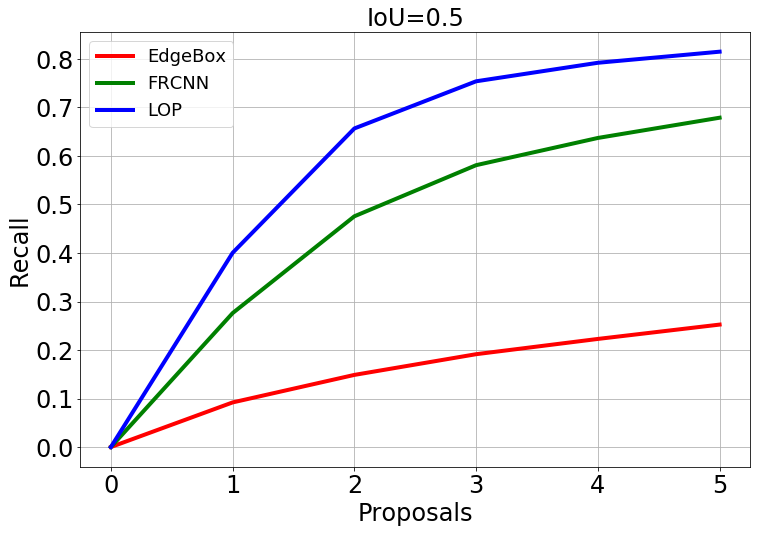}

\noindent
\includegraphics[width=0.5\linewidth]{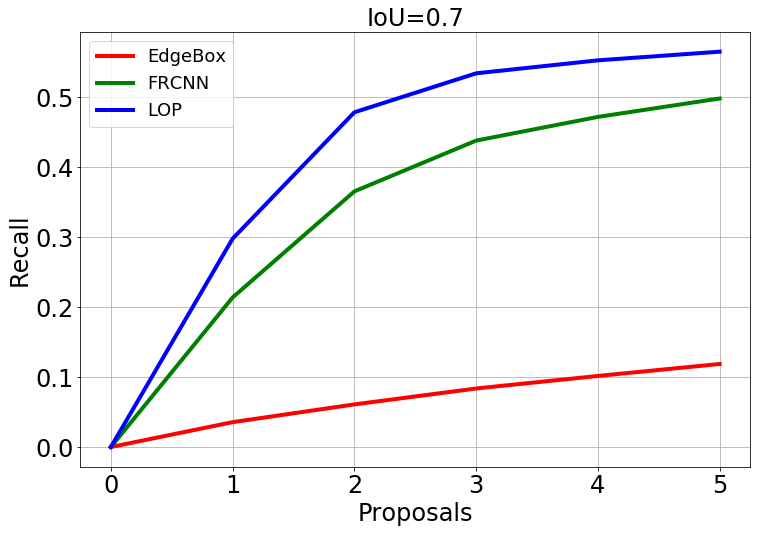}
\end{tabular}
\caption{Comparison of object proposal techniques: a) Edgebox, b) FRCNN and c) LOP (ours). Top: Recall vs IoU overlap ratio; Bottow: Recall vs Number of Proposals. The evaluation is performed on GoogleRef Validation set, which consists of 9536 objects.  }
\label{fig:recallvsiou}
\end{figure}

\begin{figure}
  \centering
  \includegraphics[width=0.45\textwidth,trim={.05\textwidth} {.00\textwidth} {0.05\textwidth} {.05\textwidth},clip]{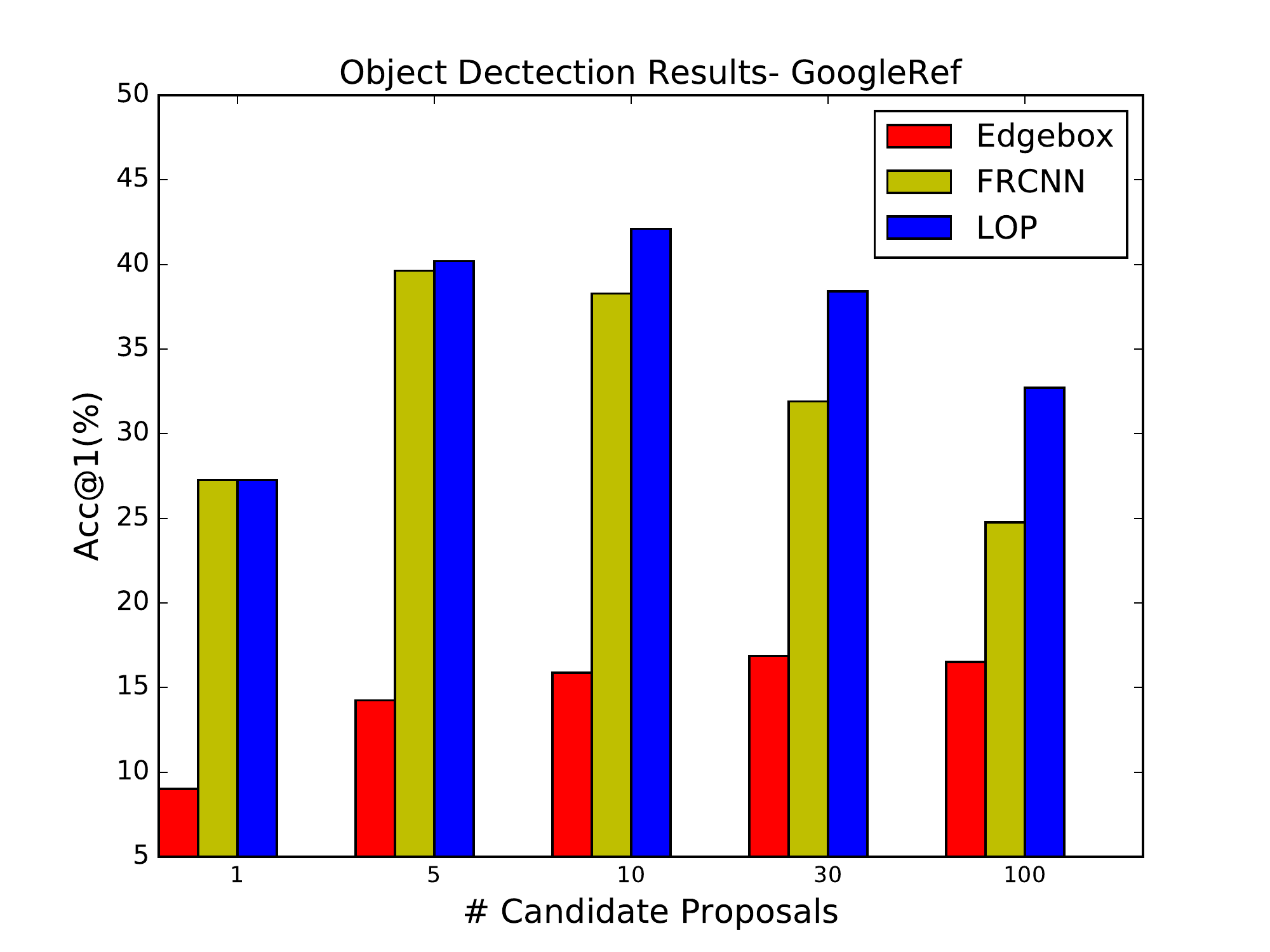}
   \caption{LID results(Acc@1) on GoogleRef dataset over the number of object proposals. The comparison is made when different object proposals technique are used: a) Edgebox, b) FRCNN and c) LOP (ours).}
  \label{fig:obj_prop_comp}
\end{figure}

Following object proposals literature\cite{ZitnickECCV14edgeBoxes, van2011segmentation}, we evaluate our LOP by the average recall of target objects under a fixed IoU criterion. It is evaluated under multiple candidate budgets. Figure~\ref{fig:recallvsiou} shows the results of LOP on GoogleRef dataset, with a comparison to EdgeBox and FRCNN which are agnostic to referring expressions. From the left column of Figure~\ref{fig:recallvsiou}, we observe that recall of LOP over all IoUs deteriorates less when we compare the plots of number of proposals being $10$ and $5$, in comparison with others. In the right column, we see that recall of LOP is high at both IoUs of $0.5$ and $0.7$ over the number of candidate proposals. The figure shows that LOP outperforms Edgebox and FRCNN consistently.

It is also interesting to see how LOP improves the performance of existing Language-grounded Instance Detection (LID) approaches. Here, we compare our LOP with Edgebox and FRCNN using the system of Hu~\textit{et al}~\cite{hu2016natural} which uses Edgebox proposal technique followed by a image captioning model. 
In Figure~\ref{fig:obj_prop_comp}, we keep the captioning model fixed and make suitable comparisons between different proposal techniques. 
We see from Figure~\ref{fig:obj_prop_comp} that LOP outperforms both EdgeBox and FRCNN detections for the task of LID. We experimented with varying numbers of object proposals = 1, 5, 10, 30 and 100. Using the $Acc@1$ metric for Object detection (given $10$ object proposals), we see that LOP improves by $3.83$\% over FRCNN detection proposals and by $26.23$\% over EdgeBox Proposals as in Figure~\ref{fig:obj_prop_comp}. The corresponding improvements of LOP over EdgeBox and FRCNN can also be observed for 5, 30 and 100 candidate proposals.
Thus, LOP keeps a check on passing poor candidate proposals to the following complex instance detection model (LID in our case). We observe the significance of LOP when we pass higher number of object proposals. In Figure~\ref{fig:obj_prop_comp}, we see clearly that difference in improvement with FRCNN increase as number of candidate proposals increases.

\section{Discussion and Conclusion}

In this work, we have proposed a solution to the problem of Object Referring in Visual Scenes with Spoken Language (ORSpoken). We have tackled the problem by presenting two datasets and a novel method. The annotated data with three modalities are ideal for multi-modality learning. This work has shown how the complicated task ORSpoken can be taken down into three subproblems, and how to develop goal-oriented vision-language interaction models for the corresponding sub-tasks. Extensive experiments show that our proposed approaches and models are superior to competing methods at all different levels: from expression transcription, to object candidate proposal, and to object referring in visual scenes with speech expressions.

We infer the following: a) when speech is relevant to the visual context, our  Visually Grounded Speech Recognition outperforms standalone speech recognition significantly; b) by using language information, our Language based Object Proposal generates in-class object proposals instead of proposals of general classes as done by previous methods like EdgeBox and FRCNN; c) the contribution of a) and b) are complementary to each other, and their combination yields a promising solution to Object Referring in Visual Scene with Spoken Language.  The aim of the work is to provide insights in this new research direction. Code and data will be made publicly available.     


\noindent
\section{Acknowledgement}
The work has been supported by Toyota via the research project TRACE-Z\"{u}rich.


{\small
\bibliographystyle{ieee}
\bibliography{sigproc}
}

\end{document}